%% file: emnlp2018.tex
\documentclass[11pt,a4paper]{article}
\usepackage[hyperref]{emnlp2018}
\usepackage{times}
\usepackage{latexsym}

\usepackage{caption}
\usepackage{graphicx}
\usepackage{float}
\usepackage{multirow}
\usepackage{amsmath}
\usepackage{url}
\usepackage{amsfonts}
\usepackage{subcaption}
\usepackage{sidecap}

\newcommand{\method}{\textsc{DDLA} }
\newcommand{\methoduni}{\textsc{DDLA-uniform} }

\usepackage{booktabs}
\usepackage[margin=1in]{geometry}
\usepackage{array}
\newcolumntype{L}[1]{>{\raggedright\arraybackslash}p{#1}}

\usepackage{url}

\aclfinalcopy 

\title{Learning to Describe Differences Between Pairs of Similar Images}

\author{Harsh Jhamtani, Taylor Berg-Kirkpatrick \\
        Language Technologies Institute \\ Carnegie Mellon University \\ {\tt \{jharsh,tberg\}@cs.cmu.edu } }

\begin{document}
\maketitle
\begin{abstract}
In this paper, we introduce the task of automatically generating text to describe the differences between two similar images. 
We collect a new dataset by crowd-sourcing difference descriptions for pairs of image frames extracted from video-surveillance footage. Annotators were asked to succinctly describe \emph{all} the differences in a short paragraph. As a result, our novel dataset provides an opportunity to explore models that 
align language and vision, and capture visual salience. The dataset may also be a useful benchmark for coherent multi-sentence generation. 
We perform 
a first-pass visual analysis that exposes clusters of differing pixels as a proxy for object-level differences. We propose a model that captures visual salience by using a latent variable to align clusters of differing pixels with output sentences. 
We find that, for both single-sentence generation and as well as multi-sentence generation, the proposed model outperforms the models that use attention alone.
\end{abstract}

\input{texfiles/intro.tex}

\input{texfiles/data.tex}

\input{texfiles/method.tex}

\input{texfiles/experiments.tex}

\input{texfiles/related.tex}

\input{texfiles/conclusions.tex}

\bibliography{emnlp2018}
\bibliographystyle{acl_natbib_nourl}

\end{document}

%% file: texfiles/intro.tex
\section{Introduction}


\begin{figure}
    \centering
    \captionsetup{font=footnotesize}
    \begin{subfigure}{\textwidth}
    \includegraphics[width=0.48\textwidth]{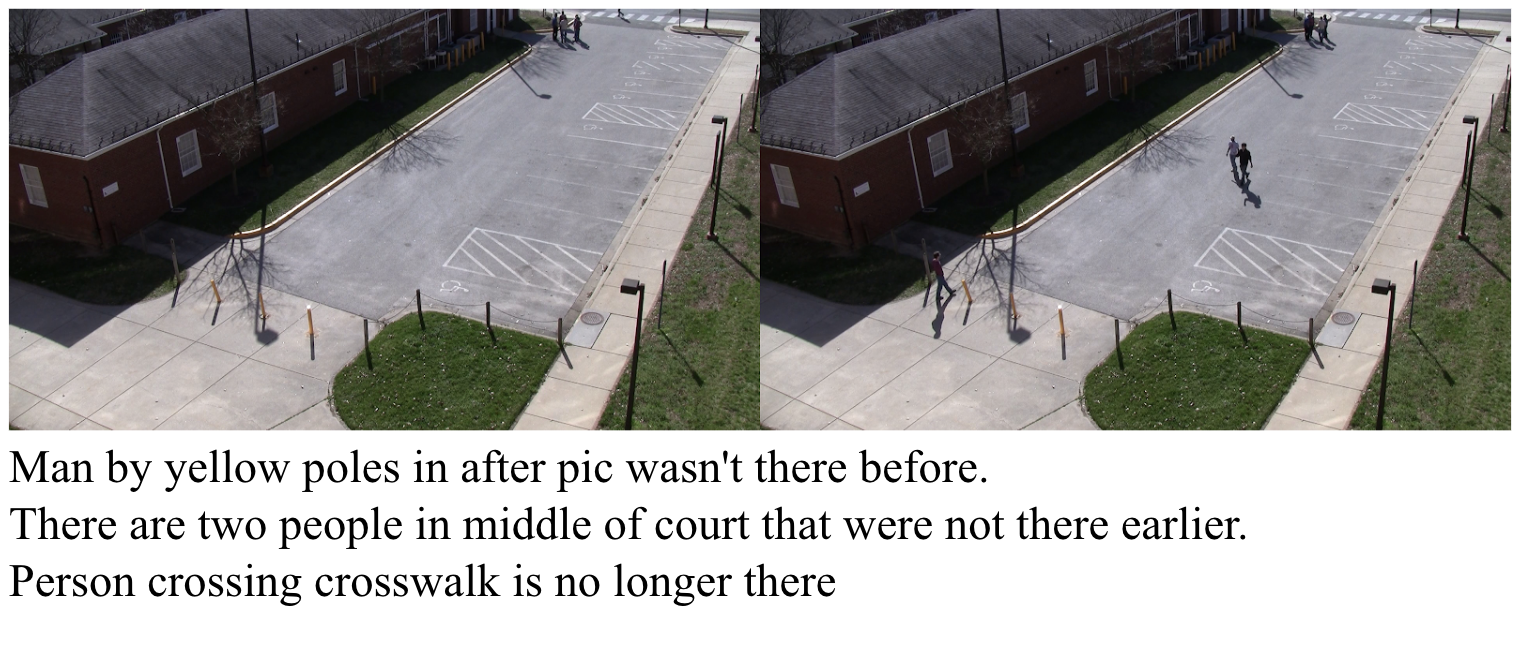}
    \end{subfigure} 
    \\ 
    \begin{subfigure}{\textwidth}
    \captionsetup{font=footnotesize}
    \includegraphics[width=0.48\textwidth]{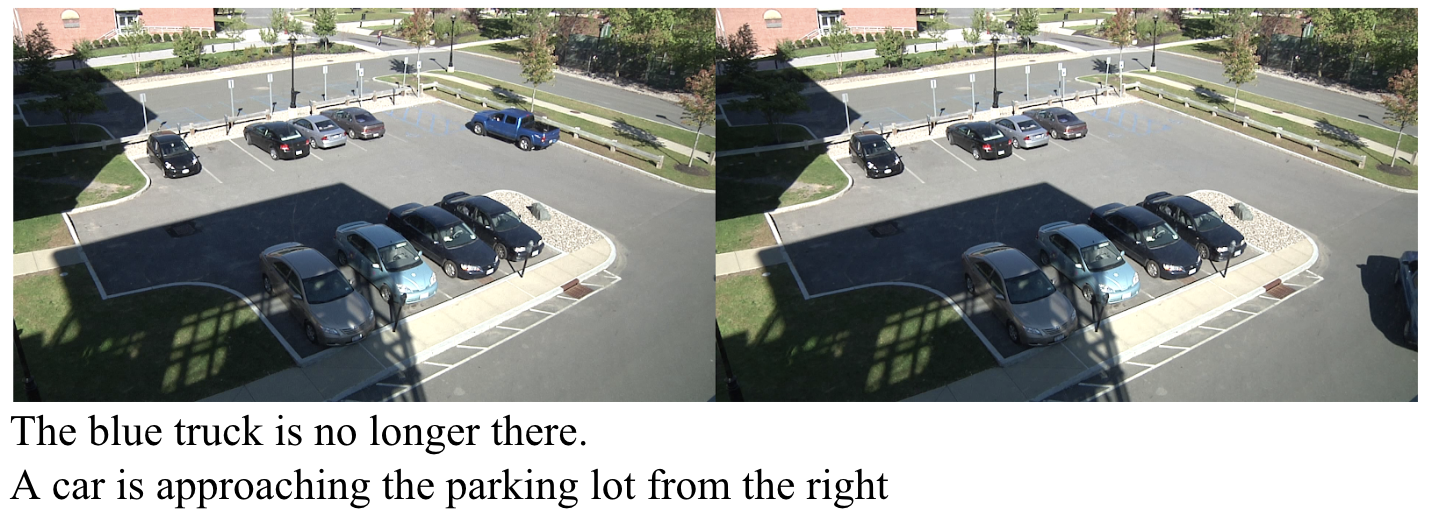}
    \end{subfigure}
    \caption{ Examples from Spot-the-diff dataset: We collect text descriptions of all the differences between a pair of images. Note that the annotations in our dataset are exhaustive wrt differences in the two images i.e. annotators were asked to describe all the visible differences. Thus, the annotations contain multi-sentence descriptions.
    }
    \label{fig:image_eg}
\end{figure}

The interface between human users and collections of data is an important application area for artificial intelligence (AI) technologies. Can we build systems that effectively interpret data and present their results concisely in natural language?
One recent goal in artificial intelligence has been to build models that are able to interpret and describe visual data to assist humans in various tasks. For example, image captioning systems \cite{vinyals2015show,xu2015show,rennie2017self, zhang2017actor} and visual question answering systems \cite{antol2015vqa,lu2016hierarchical,xu2016ask} can help visually impaired people in interacting with the world. Another way in which machines can assist humans is by identifying meaningful patterns in data, selecting and combining salient patterns, and generating concise and fluent `human-consumable' descriptions. For instance, text summarization \cite{mani1999advances,gupta2010survey,rush2015neural}
has been a long standing problem in natural language processing aimed at providing a concise text summary of a collection of documents.

In this paper, we propose a new task and accompanying dataset that combines elements of image captioning and summarization: the goal of `spot-the-diff' is to generate a succinct text description of \emph{all} the salient differences between a pair of similar images.  
Apart from being a fun puzzle, solutions to this task may have applications in assisted surveillance, as well as computer assisted tracking of changes in media assets. 
We collect and release a novel dataset for this task, which will be potentially useful for both natural language and computer vision research communities. 
We used crowd-sourcing to collect text descriptions of differences between pairs of image frames from video-surveillance footage \cite{oh2011large}, asking annotators to succinctly describe \emph{all} salient differences. In total, our datasets consist of descriptions for 13,192 image pairs.
Figure \ref{fig:image_eg} shows a sample data point - a pair of images along with a text description of the differences between the two images as per a human annotator.

There are multiple interesting modeling challenges associated with the task of generating natural language summaries of differences between images.
First, not all low-level visual differences are sufficiently salient to warrant description. The dataset presents an interesting source of supervision for methods that attempt to learn models of visual salience (we additionally conduct exploratory experiments with a baseline salience model, as described later).
Second, humans use different levels of abstraction when describing visual differences. For example, when multiple nearby objects have all moved in coordination between images in a pair, an annotator may refer to the group as a single concept (e.g. `the row of cars').
Third, given a set of salient differences, planning the order of description and generating a fluent sequence of multiple sentences is itself a challenging problem. 
Together, these aspects of the proposed task make it a useful benchmark for several directions of research.

Finally, we experiment with neural image captioning based methods. 
Since salient differences are usually described at an object-level rather than at a pixel-level, we condition these systems on a first-pass visual analysis that exposes clusters of differing pixels as a proxy for object-level differences. 
We propose a model which uses latent discrete variables in order to directly align difference clusters to output sentences. Additionally we incorporate a learned prior that models the visual salience of these  difference clusters. 
We observe that the proposed model which uses alignment as a discrete latent variable outperforms those that use attention alone. 

%% file: texfiles/data.tex
\section{`Spot-the-diff' Task and Dataset}

We introduce `spot-the-diff' dataset consisting of 13,192 image pairs along with corresponding human provided text annotations stating the differences between the two images.
Our goal was to create a dataset wherein there are meaningful differences between two similar images. To achieve this, we work with image frames extracted from VIRAT surveillance video dataset \cite{oh2011large}, which consists of 329 videos across 11 frames of reference totalling to about 8.5 hours of videos. 


\subsection{Extracting Pairs of Image Frames} 
To construct our dataset, we first need to identify image pairs such that some objects have changed positions or have entered or left in the second image compared to the first image.
To achieve this, we first extract a certain number of randomly selected image frame pairs from a given video.
Thereafter, we compute the $L_2$ distance between the two images in each pair (under RGB representation). 
Finally, we set a lower and a upper threshold on the $L_2$ distance values so calculated to filter out the image pairs with potentially too less or too many changes. These thresholds are selected based on manual inspection. 
The resulting image pairs are used for collecting the difference descriptions. 

\begin{table}[]
    \captionsetup{font=small}
    \small
    \centering
    \begin{tabular}{|c|c|}
    \hline
    \begin{tabular}{c} Total number of annotations \\  \end{tabular}  & $13,192$  \\ \hline
     \begin{tabular}{c} Mean (std dev.) number \\ of sentences per annotation \end{tabular} & $1.86 (1.01)$ \\ \hline
     \begin{tabular}{c} Vocabulary size \end{tabular} & $2404$ \\ \hline
     \begin{tabular}{c} Frequent word types \\  ($>=$5 occurrences) \end{tabular} & $1000$ \\ \hline
     \begin{tabular}{c} Word tokens that are \\ frequent word types \end{tabular} & $97\%$ \\ \hline
     \begin{tabular}{c} Mean (std dev.) number \\ of words in sentence: \end{tabular} &  $10.96 (4.97)$ \\ \hline
     \begin{tabular}{c} \% Long sentences \\ ($>20$ words) \end{tabular} & $5\%$ \\ \hline
    \end{tabular}\caption{Summary statistics for spot-the-diff dataset}
    \label{table:data}
\end{table}

\subsection{Human Annotation}

\begin{SCfigure*}[]
    \captionsetup{font=small}
    \centering
    \includegraphics[width=0.76\textwidth]{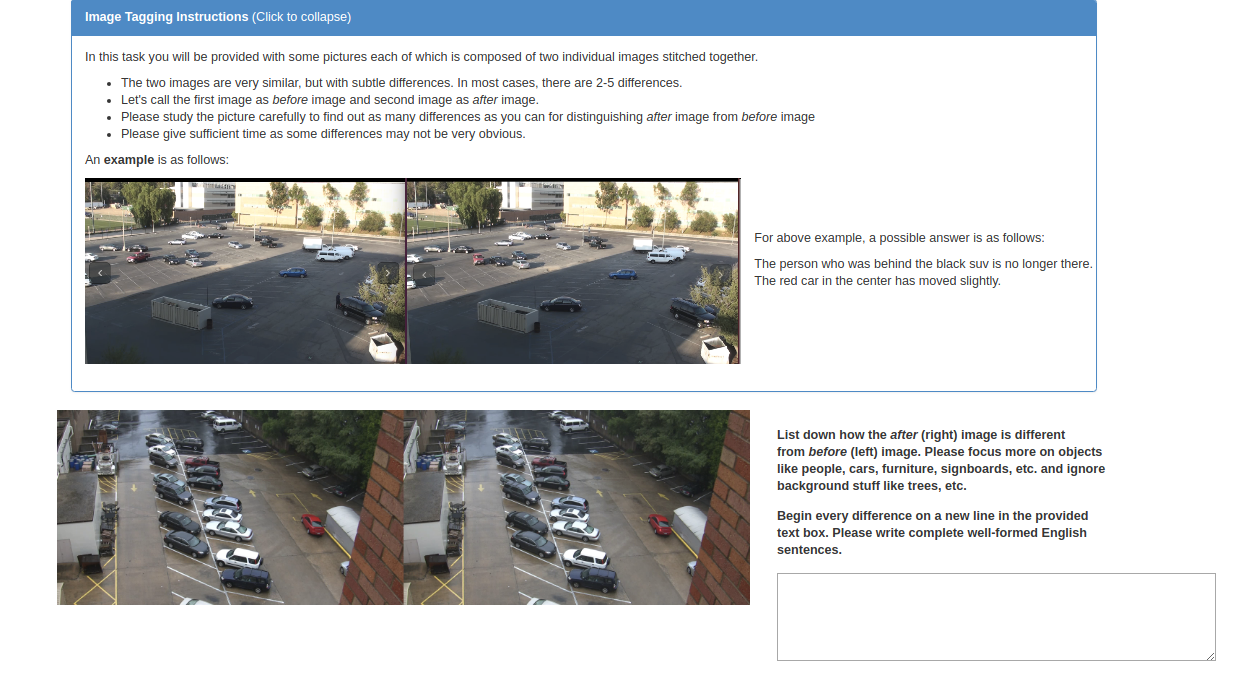}
    \caption{AMT (Amazon Mechanical Turk) HIT (Human Intelligence Task) setup for data collection. We provide the annotators with detailed instructions, along with an example showing how to perform the task. 
    We request the annotators to write complete English sentences, with each sentence on a separate line. We collect a total of 13,192 annotations.
    }
    \label{fig:amt_setup}
\end{SCfigure*}

We crowd-sourced natural language differences between images using Amazon Mechanical Turk. 
We restrict to annotators from primarily Anglophone countries: USA, Australia, United Kingdom, and Canada, as we are working with English language annotations. We limit to those participants which have lifetime HIT $>80\%$. We award 5 cents per HIT (Human Intelligence Task) to participants.
We provide the annotators with an example on how to work on the task. We request the annotators to write complete English sentences, with each sentence on a separate line. We collect a total of 13192 annotations. 


\begin{table}[]
    \centering
    \small
    \begin{tabular}{ccc}
        \begin{tabular}{c} \textbf{Dataset}  \end{tabular} & \textbf{BLEU-1/2/3/4} & \textbf{ROUGE-L}  \\ \hline
         \begin{tabular}{c} Spot-the-diff \\ ($A=3$) \end{tabular} & $0.41/0.25/0.15/0.08$ &  $0.31$ \\ \hline
        \begin{tabular}{c} MS-COCO \\ ($A=3$) \end{tabular}  & $0.38/0.22/0.13/0.08 $ & $0.34$ \\  \hline
        \begin{tabular}{c} MS-COCO \\ ($A=5$) \end{tabular} & $0.66/0.47/0.32/0.22$ & $0.48$ \\
          \hline
    \end{tabular}
\caption{ \footnotesize Human agreement for our dataset: We report measures such as BLEU and ROUGE when `evaluating' one set of human generated
captions against the remaining sets. $A=k$ represents $k$ captions per data point, out of which 1 is chosen as hypothesis, while remaining $k-1$ act as references.
}
\label{table:agreement}
\end{table}

\subsection{Dataset statistics}
Table \ref{table:data} shows some summary statistics about the collected dataset. Since we deal with a focused domain, we observe a small vocabulary size.   On an average there are 1.86 reported differences / sentences per image pair.
We also report inter-annotator agreement as measured using text overlap of multiple annotations for the same image pair. We collect three sets of annotations for a small subset of the data (467 data points) for the purpose of reporting inter-annotator agreements. We thereby calculate BLEU and ROUGE-L scores by treating one set of annotations as `hypothesis' while remaining two sets act as `references'(Table \ref{table:agreement}). 
We repeat the same analysis for MS-COCO dataset and report these measures for reference. 
The BLEU and METEOR values for our dataset seem reasonable and are comparable to the values observed for MS-COCO dataset.

%% file: texfiles/method.tex

\section{Modeling Difference Description Generation}
We propose a neural model for describing visual difference based on the input pair of images that uses latent alignment variable to capture visual salience. Since most descriptions talk about higher-level differences rather than individual pixels, we first perform a visual analysis that pre-computes a set of difference clusters in order to approximate object-level differences, as described next. The output of this analysis is treated as input to a neural encoder-decoder text generation model that incorporates a latent alignment variable and is trained on our new dataset. 

\subsection{Exposing Object-level Differences}

We first analyze the input image pair for the pixel-level differences by computing a \emph{pixel-difference mask}, followed by a local spatial analysis which segments the difference mask into clusters that approximate the set of object-level differences. 
Thereafter, we extract image features using convolutional neural models and use these as input to a neural text generation model, described later. 
\\

\noindent \textbf{Pixel-level analysis:}
The lowest level of visual difference is individual differences between corresponding pixels in the input pair. Instead of requiring our description model to learn to compute pixel-level differences as a first step, we pre-compute and directly expose these to the model.
Let $X=(\boldsymbol{I_1}, \boldsymbol{I_2})$ represent the image pair in a datum.
For each such image pair in our dataset, we obtain a corresponding pixel-difference mask $\boldsymbol{M}$. $M$ is a \emph{binary-valued} matrix of the same dimensions (length and width) as each of the images in the corresponding image pair, wherein each element in the matrix is 1 (active) if the corresponding pixel is \emph{different} between the input pair, and 0 otherwise. To decide whether a pair of corresponding pixels in the input image pair are sufficiently \emph{different}, we calculate the  $L_2$-distance between the vectors corresponding to each pixel's color value (three channels) and check whether this difference is greater than a threshold $\delta$ (set based on manual inspections). 

While the images are extracted from supposedly still cameras, we do find some minor shifts in the camera alignment, which is probably due to occasional wind but may also be due to manual human interventions. These shifts are rare and small, and we align the images in the pair by iterating over a small range of vertical and horizontal shifts to find the shift with minimum corresponding $L_2$-distance between the two images. \\

\begin{figure}
    \captionsetup{font=footnotesize}
    \centering
    \includegraphics[width=0.42\textwidth]{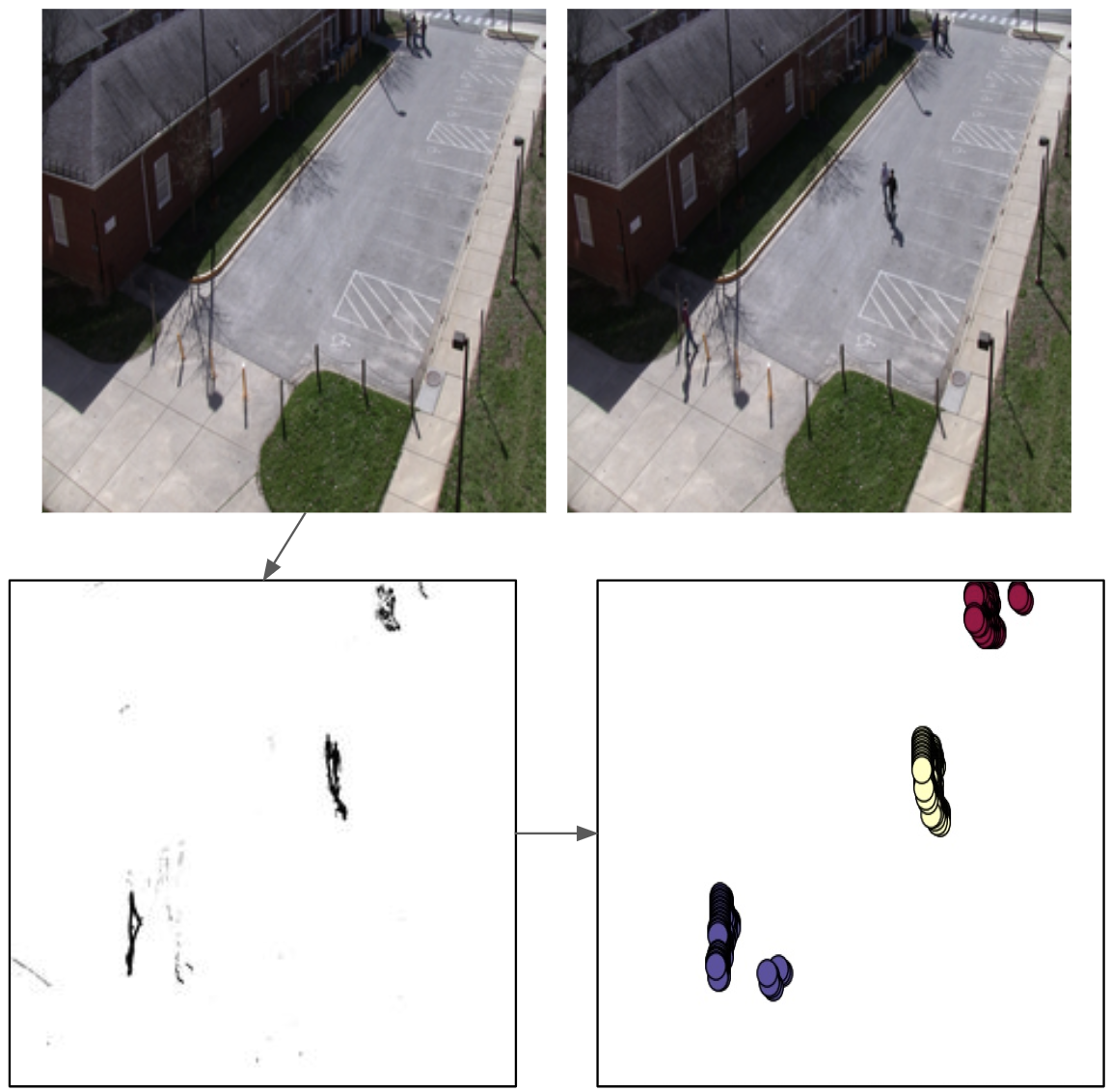}
    \caption{
    Exposing Object-level Differences: Before training a model to describe visual difference, we first compute pixel-level differences, as well as a segmentation of these differences into clusters, as a proxy for exposing object-level differences. The first row shows the original image pair. Bottom left depicts the pixel-difference mask, which represents extracted pixel-level differences. The segmentation of the pixel-difference mask into clusters is shown in the bottom right.
    }
    \label{fig:clustering}
\end{figure}

\noindent \textbf{Object-level analysis:}
Most visual descriptions refer to object-level differences rather than pixel-level differences. Again, rather than requiring the model to learn to group pixel differences into objects, we attempt to expose this to the model via pre-processing. As a proxy for object-level difference, we segment the pixel-level differences in the pixel-difference mask into clusters, and pass these clusters as additional inputs to the model. Based on manual inspection, we find that with the right clustering technique, this process results in groupings that roughly correspond to objects that have moved, appeared, and disappeared between the input pair. 
Here, we find that density based clustering algorithms like DBScan \cite{ester1996density} work well in practice for this purpose. In our scenario, the DBScan algorithm predicts clusters of nearby active pixels, and marks outliers consisting of small groups of isolated active pixels, based on a calculation of local density. This also serves as a method for pruning any noisy pixel differences which may have passed through the pixel-level analysis. 

As the output of DBScan, we obtain segmentation of the pixel difference matrix $M$ into \emph{difference clusters}. 
Let the number of \emph{difference clusters} be represented by $K$ (DBScan is a non-parametric clustering method, and as such the number of clusters $K$ is different for each data point.).
Now, let's define $\boldsymbol{C_k}$ as another binary-valued mask matrix such that the elements in matrix corresponding to the $k^{th}$ difference cluster are 1 (active) while rest of the elements are 0. 

\begin{figure}
    \captionsetup{font=footnotesize}
    \centering
    \includegraphics[width=0.48\textwidth]{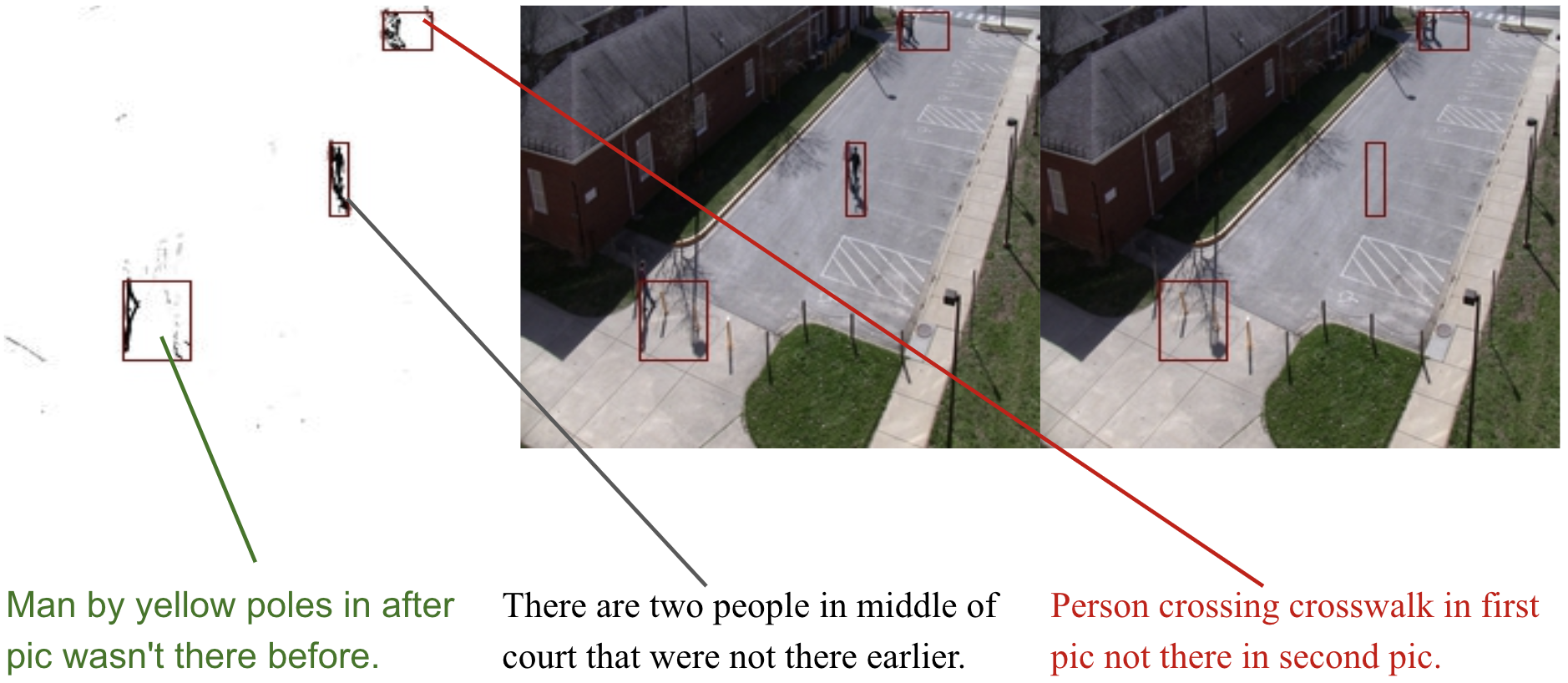}
    \caption{
   The figure shows the \emph{pixel-difference} mask for the running example, along with the two original images, with bounding boxes around clusters.
   Typically one or more difference clusters are used to frame one reported difference / sentence, and it is rare for a difference cluster to participate in more than one reported difference.
    }
    \label{fig:alignment}
\end{figure}


\subsection{Text Generation Model}

We observe from annotated data that each individual sentence in a full description typically refers only to visual differences within a single cluster (see Figure~\ref{fig:alignment}). Further, on average, there are more clusters than there are sentences. While many uninteresting and \emph{noisy} pixel-level differences get screened out in preprocessing, some uninteresting clusters are still identified. These are unlikely to be described by annotators because, even though they correspond to legitimate visual differences, they are not visually salient. Thus, we can roughly model description generation as a cluster selection process.

\def\arraystretch{1.05}
\begin{table}[t]
  \centering
  \small
  \begin{tabular}{|lcl|}
    \hline
    $X$$=$($I_1$,$I_2$) &: &Image pair in the datum \\[0.1em]
    $M$ &: &Pixel-difference mask is a binary-valued \\ && matrix depicting pixel-level changes \\[0.1em]
    $F_1, F_2$ &: &Image feature tensors for $I_1$ and $I_2$ \\ && respectively \\[0.1em]
    $K$ &: &Number of segments \\[0.1em]
    $C_k$ &: &Cluster mask corresponding to $k^{th}$ \\ && difference cluster \\[0.1em]
    $T$ &: &Number of reported differences / \\ && sentences \\[0.1em]
    $z_i$ &: &Discrete alignment variable for the $i^{th}$ \\ && sentence. $z_i \in \{1,2,...,K\}$ \\[0.1em]
    $S_1,..,S_{T}$ &: &List of T Sentences \\ \hline
  \end{tabular}
  \caption{Summary of notation used in description of the method.}
  \label{tab:notation}
\end{table}
\def\arraystretch{1}

\begin{figure*}
    \captionsetup{font=footnotesize}
    \centering
    \includegraphics[width=0.66\textwidth]{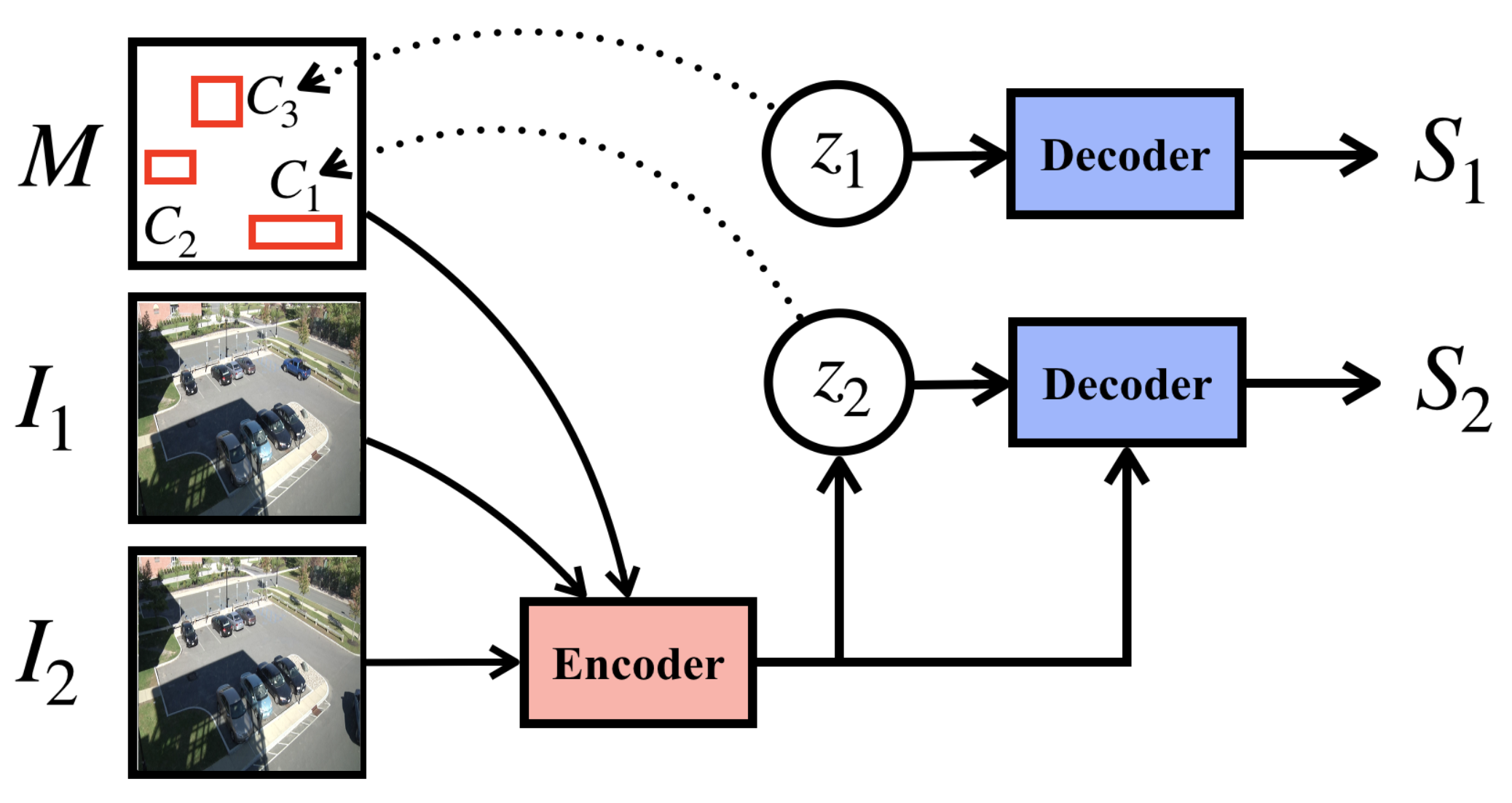}
    \caption{Model architecture for generating difference descriptions. We incorporate a discrete latent variable $z$ which selects one of the clusters as a proxy for object-level focus. Conditioned on the cluster and visual features in the corresponding region, the model generates a sentence using an LSTM decoder. During training, each sentence in the full description receives its own latent alignment variable, $z$. }
    \label{fig:arch}
\end{figure*}

In our model, which is depicted in Figure~\ref{fig:arch}, we assume that each output description, which consists of sentences $S_1, \ldots,  S_T$, is generated sentence by sentence conditioned on the input image pair $X=(I_1,I_2)$. Further, we let each sentence $S_i$ be associated with a latent alignment variable, $z_i \in \{1, \ldots, K\}$, that chooses a cluster to focus on \cite{vinyals2015pointer}. The choice of $z_i$ is itself conditioned on the input image pair, and parameterized in a way that lets the model learn which types of clusters are visually salient and therefore likely to be described as sentences. Together, the probability of a description given an image pair is given by:
\begin{multline}
    \label{eqn:marg}
     P(S_1,..,S_T|X) \\ = \sum_{z_1,..,z_T} \prod_{i=1}^{T} \underbrace{P(S_i|z_i,X;\theta))}_\text{decoder} \underbrace{P(z_i|X; w)}_\text{alignment prior}  
\end{multline}
The various components of this equation are described in detail in the next few sections. Here, we briefly summarize each. The term $P(z_i|X;w)$ represents the prior over the latent variable $z_i$ and is parameterized in a way that lets the model learn which types of clusters are visually salient. The term $P(S_i|z_i,X;\theta)$ represents the likelihood of sentence $S_i$ given the input image pair and alignment $z_i$. We employ masking and attention mechanisms to encourage this decoder to focus on the cluster chosen by $z_i$. Each of these components conditions on visual features produced by a pre-trained image encoder. 

The alignment variable $z_i$ for each sentence is chosen independently, and thus our model is similar to IBM Model 1 \cite{brown1993mathematics} in terms of its factorization structure. This will allow tractable learning and inference as described in Section~\ref{ssec:learning}. We refer to our approach as \method  
(\textbf{D}ifference \textbf{D}escription with \textbf{L}atent \textbf{A}lignment). \\


\noindent \textbf{Alignment prior:}
We define a learnable prior over alignment variable $z_i$. In particular, we let the multinomial distribution on $z_i$ be parameterized in a log-linear fashion using feature function $g(z_i)$. Specifically, we consider the following four features: the length, width, and area of the smallest rectangular region enclosing cluster $z_i$, and the number of active elements in mask $C_{z_i}$. Specifically, we let
    $P(z_i|X;w) \propto \exp(w^T g(z_i))$. \\

\noindent \textbf{Visual encoder:}
We extract images features using ResNet \cite{he2016deep} pre-trained on Imagenet data. Similar to prior work \cite{xu2015show}, we extract features using a lower level convolutional layer instead of fully connected layer.
In this way, we obtain image features of dimensionality $14*14*2096$, where the first two dimensions correspond to a grid of coarse, spatially localized, feature vectors. Let $F_1$ and $F_2$ represent the extracted feature tensors for $I_1$ and $I_2$ respectively. 
\\

\noindent \textbf{Sentence decoder:}
We use an LSTM decoder \cite{hochreiter1997long} to generate the sequence of words in each output sentence, conditioned on the image pair and latent alignments. We use a matrix transformation of the extracted image features to initialize the hidden state of the LSTM decoder for each sentence, independent of the setting of $z_i$. Additionally, we use an attention mechanism over the image features at every decoding step, 
similar to the previous work \cite{xu2015show}. 
However, instead of considering attention over the entire image, we restrict attention over image features to the cluster mask determined by the alignment variable,  $C_{z_i}$.
Specifically, we project binary mask $C_{z_i}$ from the input image dimensionality (224*224) to the dimensionality of the visual features (14*14).
 To achieve this, we use pyramid reduce down-sampling on a smoothed version of cluster mask $C_{z_i}$. 
The resulting projection roughly corresponds to the subset of visual features with the cluster region in their receptive field. This projection is multiplied to attention weights. 


\subsection{Learning and Decoding \label{ssec:learning}}

Learning in our model is accomplished by stochastic gradient ascent on the marginal likelihood of each description with alignment variables marginalized out. 
Since alignment variables are independent of one another, we can marginalize over each $z_i$ separately. This means running backpropagation through the decoder $K$ times for each sentence, where $K$ is the number of clusters. In practice $K$ is relatively small and this direct approach to training is feasible. 
Following equation \ref{eqn:marg}, we train both the generation and prior in an end-to-end fashion.

For decoding, we consider the following two problem settings. In the first setting, we consider the task of producing a single sentence in isolation. We evaluate in this setting by treating the sentences in the ground truth description as multiple reference captions. This setting is similar to the typical image captioning setting. In the second setting, we consider the full multi-sentence generation task where the system is required to produce a full description consisting of multiple sentences describing all differences in the input. Here, the generated multi-sentence text is  directly evaluated against the multi-sentence annotation in the crowd-soured data. \\

\noindent \textbf{Single-sentence decoding:} 
For single sentence generation, 
we first select the value of $z_i$ which maximizes the prior $P(z_i|X;w)$. Thereafter, we simply use greedy decoding to generate a sentence conditioned on the chosen $z_i$ and the input image pair. \\

\noindent \textbf{Multi-sentence decoding:} 
Here, we first select a set of clusters to include in the output description, and then generate a single sentence for each cluster using greedy decoding. Since typically there are more clusters than sentences, we condition on the ground truth number of sentences and choose the corresponding number of clusters.
We rank clusters by decreasing likelihood under the alignment prior and then choose the top $T$.  



%% file: texfiles/experiments.tex

\section{Experiments}

\begin{table*}[ht]
\small
\begin{center}
\begin{tabular}{c c c c c c c}
\hline \textbf{Model} & \textbf{Bleu 1/2/3/4 }& \textbf{Meteor} & \textbf{Cider} & \textbf{Rouge-L} & \textbf{Perplexity} \\ \hline
\begin{tabular}{c} \textsc{NN}  \end{tabular} & 0.226 0.111 0.057 0.026 & 0.102  & 0.120  & 0.201 & -  \\ 
\begin{tabular}{c}\textsc{CAPT} \end{tabular} & 0.304 0.194 0.126 0.073 & 0.105  & 0.263 & 0.256 & 16.78  \\
\begin{tabular}{c}\textsc{CAPT-masked} \end{tabular} & 0.301 0.200 0.131 0.078 & 0.108  &  0.285 & 0.271  & 15.12  \\
\begin{tabular}{c} \methoduni \end{tabular} & 0.285 0.175 0.108 0.064  & 0.106  & 0.250 & 0.247 & 9.96  \\
\begin{tabular}{c} \method \end{tabular} & 0.343 0.221 0.140 0.085  & 0.120  & \textbf{0.328} & 0.286 & 9.73  \\ 
\hline
\end{tabular}
\end{center}
\caption{\footnotesize \label{table:base01}  \textbf{Single sentence decoding:} We report automatic evaluation scores for various models under single sentence generation setting. \method model fares better scores than various baseline methods for all the considered measures. Both the \method models get much better perplexities than baseline methods. }
\end{table*}

\begin{table*}[t]
\small
\begin{center}
\begin{tabular}{c c c c c c}
\hline \textbf{Model} & \textbf{Bleu 1/2/3/4 }& \textbf{Meteor} & \textbf{Cider} & \textbf{Rouge-L} & \textbf{LenRatio} \\ \hline
\begin{tabular}{c}\textsc{NN-multi}\end{tabular} & 0.223 0.109 0.056 0.026 & 0.087  & 0.105  & 0.181  & 1.035   \\
\begin{tabular}{c}\textsc{CAPT-multi} \end{tabular} & 0.262 0.146 0.081 0.045 & 0.094  & 0.235  & 0.174 &  1.042   \\
\begin{tabular}{c}\methoduni \end{tabular} & 0.243 0.143 0.085 0.051 & 0.094 & 0.217 & 0.213 & 0.778 \\
\begin{tabular}{c}\method \end{tabular} & 0.289 0.173 0.103 0.062  &  0.108 &  \textbf{0.297} & 0.260 & 0.811 \\
\hline
\end{tabular}
\end{center}
\caption{\footnotesize \label{table:para} \textbf{Multi-sentence decoding} We report automatic evaluation scores for various models under multi-sentence generation setting. \method model achieves better scores compared to the baseline methods. Note that these scores are not directly comparable with single sentence generation setting. \textbf{LenRatio} is the ratio of the average number of tokens in the prediction to the average number of tokens in the ground truth for the test set. }
\end{table*}



We split videos used to create the dataset into train, test, and validation in the ratio 80:10:10. This is done to ensure that all data points using images from the same video are entirely in one split. 
We report quantitative metrics like CIDEr \cite{vedantam2015cider}, BLEU \cite{papineni2002bleu}, METEOR \cite{denkowski2014meteor}, and ROUGE-L, as is often reported by works in image captioning.
We report these measures for both sentence level setting and multi-sentence generation settings. Thereafter, we also discuss some qualitative examples.
We implement our models in PyTorch \cite{paszke2017automatic}. 
We use mini-batches of size 8 and use Adam optimizer
\footnote{Our data set can be obtained through \url{https://github.com/harsh19/spot-the-diff} }. We use CIDEr scores on validation set as a criteria for early stopping. \\

\noindent \textbf{Baseline models:}
We consider following baseline models: \textsc{Capt} model considers soft attention over the input pair of images (This attention mechanism is similar to that used in prior image captioning works \cite{xu2015show}, except that we have two images instead of a single image input). We do not perform any masking in case of \textsc{Capt} model, and simply ignore the cluster information. The model is trained to generate a single sentence. Thus, this model is similar to a typical captioning model but with soft attention over two images.
\textsc{Capt-mask} model is similar to \textsc{Capt} model except that it incorporates the masking mechanism defined earlier using the union of all the cluster masks in the corresponding image.
We also consider a version of the \textsc{Capt} model wherein the target prediction is the whole multi-sentence description -- \textsc{Capt-multi} -- for this setting, we simply concatenate the sentences in any arbitrary order \footnote{Note that we do not provide \textsc{Capt-multi} with ground truth number of sentences}.  Additionally, we consider a nearest neighbor baseline (\textsc{NN-multi}), wherein we simply use the annotation of the closest matching training data point. We compute the closeness based on the extracted features of the image pair, and leverage  sklearn’s \cite{pedregosa2011scikit} Nearest-Neighbor module. For single sentence setting (\textsc{NN}), we randomly pick one of the sentences in the annotation. 

We also consider a version of \method model with fixed uniform prior, and refer to this model as \methoduni. For single sentence generation, we sample $z_j$ randomly from the uniform distribution and then perform decoding. For the multi-sentence generation setting, we employ simple heuristics to order the clusters at test time. One such heuristic we consider is to order the clusters as per the decreasing area of the bounding box (smallest rectangular area enclosing the cluster). 
\\

\begin{figure*}
    \captionsetup{font=footnotesize}
    \centering
    \includegraphics[width=0.87\textwidth]{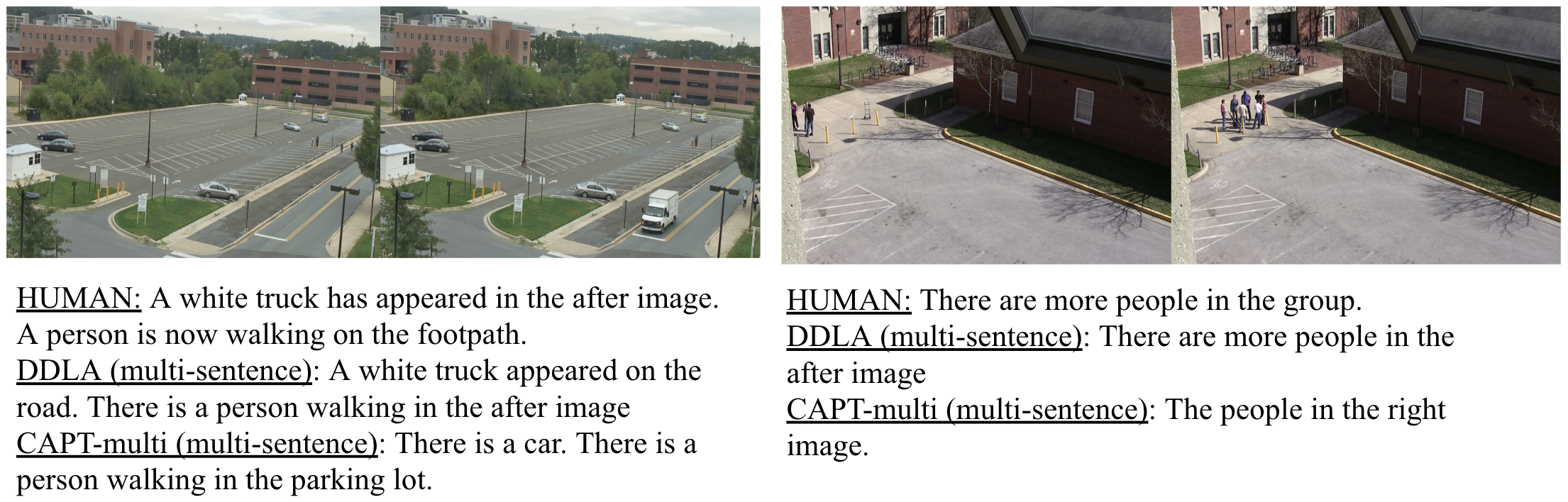}
    \caption{Predictions from various methods for two input image pairs.}
    \label{fig:output}
\end{figure*}

\noindent \textbf{Results:}
We report various automated metrics for the different methods under single sentence generation and multi-sentence generation in Tables \ref{table:base01} and \ref{table:para} respectively.
For the single sentence generation setting, we observe that the \method model outperforms various baselines as per most of the scores on the test data split. \methoduni method performs similar to the \textsc{Capt} baseline methods. 
For the multi-sentence generation, the \method model again outperforms other methods. This means that having a learned prior is useful in our proposed method. 
Figure \ref{fig:output} shows an example data point with predicted outputs by different methods.


\section{Discussion and Analysis}

\begin{figure}
    \captionsetup{font=footnotesize}
    \centering
    \includegraphics[width=0.48\textwidth]{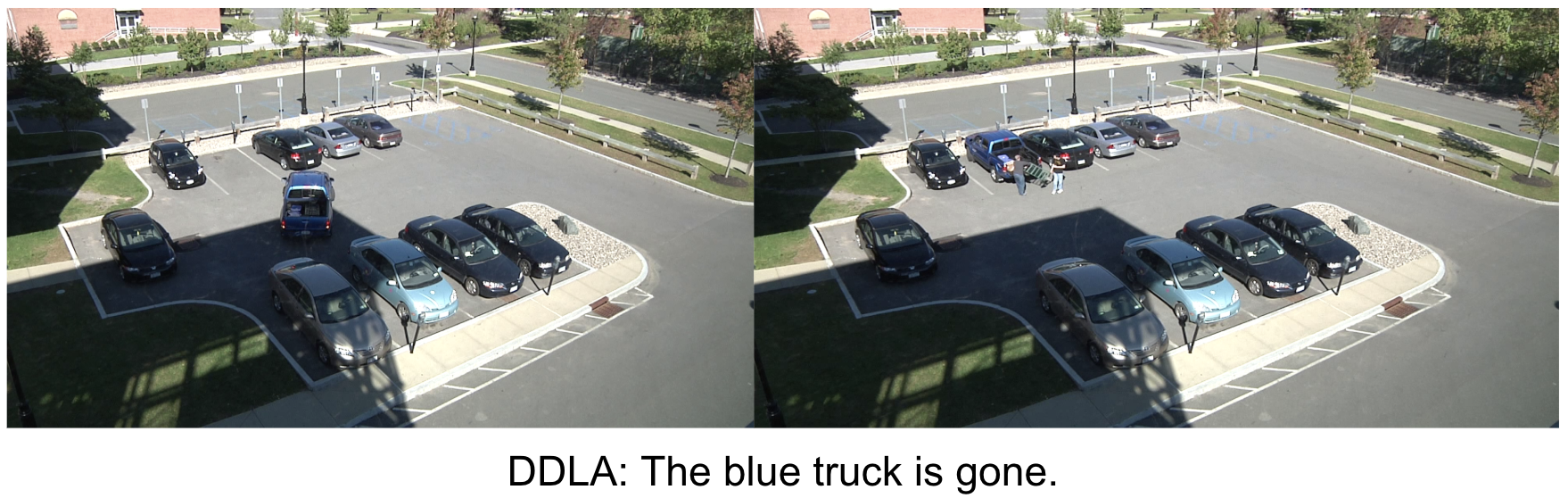}
    \caption{Some drawbacks with the current models: One apparent drawback with the single cluster selection is that it misses opportunity to identify an object which has moved significantly- considering it as appeared or disappeared as the case may be. In this example, the blue truck moved, but the \method model predicts that the truck is no longer there.}
    \label{fig:issue}
\end{figure}

\noindent \textbf{Qualitative Analysis of Outputs}
We perform a qualitative analysis on the outputs to understand the drawbacks in the current methods. One apparent limitation of the current methods is the failure to explicitly model the movement of same object in the two images (Figure \ref{fig:issue}) -- prior works on object tracking can be useful here. 
Sometimes the models get certain attributes of the objects wrong. e.g. `blue car' instead of `red car'.
Some output predictions state an object to have `appeared' instead of `disappeared' and vice versa.  \\ 

\noindent \textbf{Do models learn alignment between sentence and difference clusters?}
We performed a study on 50 image pairs by having two humans manually annotate gold alignments between sentences and difference clusters. We then computed alignment precision for the model's predicted alignments. To obtain model's predicted alignment for a given sentence $S_i$, we compute $argmax_k  P(z_i=k|X)P(S_i|z_i=k,X)$.
Our proposed model achieved a precision of 54.6\%, an improvement over random chance at 27.4\%. \\ 

\noindent \textbf{Clustering for pre-processing}
Our generation algorithm assumed one sentence uses only one cluster and as such we tune the hyper-parameters of clustering method to get large clusters so that typically a cluster will entirely contain a reported difference. On inspecting randomly selected data points, we observe that in some cases too large clusters are marked by the clustering procedure. One way to mitigate this is to tune clustering parameters to get smaller clusters and update the generation part to use a subset of clusters. 
As mentioned earlier, we consider clustering as a means to achieve object level pre-processing. One possible future direction is to leverage pre-trained object detection models to detect cars, trucks, people, etc. and make these predictions readily available to the generation model. \\

\noindent \textbf{Multi-sentence Training and Decoding}
As mentioned previously, we query the models for a desired number of 'sentences'. In future works we would like to relax this assumption and design models which can predict the number of sentences as well. Additionally, our proposed model doesn't not explicitly ensure consistency in the latent variables for different sentences of a given data point i.e the model does not make explicit use of the fact that sentences report non-overlapping visual differences. Enforcing this knowledge while retaining the feasibility of training is a potential future direction of work.

%% file: texfiles/related.tex
\section{Related Work}

 
\noindent \textbf{Modeling pragmatics:} 
The dataset presents an opportunity to test methods which can model pragmatics and reason about semantic, spatial and visual similarity to generate a textual description of what has changed from one image to another. 
Some prior work in this direction \cite{andreas2016reasoning, vedantam2017context} contrastively describe a target scene in presence of a distractor. 
In another related task -- referring expression comprehension \cite{kazemzadeh2014referitgame,mao2016generation,hu2017modeling} -- the model has to identify which object in the image is being referred to by the given sentence. 
However, our proposed task comes with a pragmatic goal related to summarization: the goal is to identify and describe \emph{all} the differences. Since the goal is well defined, it may be used to constrain models that attempt to learn how humans describe visual difference.  \\

\noindent \textbf{Natural language generation:}
Natural language generation (NLG) has a rich history of previous work, including, for example, recent works on biography generation \cite{lebret2016neural}, weather report generation \cite{mei2015talk}, and recipe generation \cite{kiddon2016globally}. 
Our task can viewed as a potential benchmark for coherent multi-sentence text generation since it involves assembling multiple sentences to succinctly cover a set of differences. \\ 

\noindent \textbf{Visual grounding:} 
Our dataset may also provide a useful benchmark for training unsupervised and semi-supervised models that learn to align vision and language. 
\citet{plummer2015flickr} 
collected annotation for phrase-region alignment in an image captioning dataset, 
and follow up work has attempted to predict these alignments \cite{wang2016learning,plummer2017phrase,rohrbach2016grounding}. Our proposed dataset poses a related alignment problem: attempting to align sentences or phrases to visual differences. 
However, since differences are contextual and depend on visual comparison, our new task may represent a more challenging scenario as modeling techniques advance. \\

\noindent \textbf{Image change detection:} 
There are some works on land use pattern change detection (\cite{radke2005image}). These works are related since they try to screen out noise and mark the regions of change between two images of same area at different time stamps. 
\newcite{bruzzone2000automatic} propose an unsupervised change detection algorithms aim to discriminate between changed and unchanged pixels for multi-temporal remote sensing images. \newcite{zanetti2016generalized} propose a method that allows unchanged class to be more complex rather than having a single unchanged class. 
Though image diff detection is part of our pipeline, our end task is to generate natural language descriptors. 
Moreover, we observe that simple clustering seems to work well for our dataset. \\

\noindent \textbf{Other relevant works:}
\newcite{maji2012discovering} aim to construct a lexicon of parts and attributes by formulating an annotation task where annotators are asked to describe differences between two images.  
Some other related works model phrases describing change in color \cite{winn2018lighter}, move-by-move game commentary for describing change in game state \cite{jhamtani2018learning}, and code commit message summarizing changes in code-base from one commit to another \cite{jiang2017automatically}.
There exist some prior works on fine grained image classification and captioning \cite{wah2014similarity,Nilsback06,khosla2011novel}. The premise of such works is that it is difficult for machine to find discriminative features between similar objects e.g. birds of different species. Such works are relevant for us as the type of data we deal with are usually of same object or scene taken at a different time or conditions.

%% file: texfiles/conclusions.tex
\section{Conclusion}
In this paper, we proposed the new task of describing differences between pairs of similar images and introduced a corresponding dataset. 
Compared to many prior image captioning datasets, text descriptions in the `Spot-the-diff' dataset are often multi-sentence, consisting of all the differences in two similar images in most of the cases. We performed exploratory analysis of the dataset and highlighted potential research challenges. We discuss how our 'Spot-the-diff' dataset is useful for tasks such as language vision alignment, referring expression comprehension, and multi-sentence generation.
We performed pixel and object level preprocessing on the images to identify clusters of differing pixels. We observe that the proposed model which aligns clusters of differing pixels to output sentences performs better than the models which use attention alone. 
We also discuss some limitations of current methods and scope for future directions.


\section*{Acknowledgements} 
We are thankful to anonymous EMNLP reviewers for their valuable suggestions. We thank Eric Nyberg for discussions on dataset collection. We also acknowledge Nikita Duseja and Varun Gangal for helping with the proof-reading of the paper. We thank \newcite{Luo2017} for releasing a PyTorch implementation of many popular image captioning models. This project was supported in part by a Adobe Research gift. Opinions and findings in this paper are of the authors, and do not necessarily reflect the views of Adobe.